%% file: main.tex
\documentclass[10pt,twocolumn,letterpaper]{article}

\usepackage[pagenumbers]{cvpr} 


\definecolor{cvprblue}{rgb}{0.21,0.49,0.74}
\usepackage[pagebackref,breaklinks,colorlinks,allcolors=cvprblue]{hyperref}
\usepackage{graphicx} 
\usepackage{multicol}
\usepackage{multirow}
\usepackage{booktabs}
\usepackage{amsfonts}
\usepackage{amsmath}
\usepackage{amssymb}
\usepackage{url}
\usepackage[capitalize]{cleveref}
\usepackage{fnpct}
\usepackage{caption}
\usepackage{subcaption}
\usepackage{algorithm}
\usepackage{algpseudocode}
\input{math_commands}

\def\paperID{12677} 
\def\confName{CVPR}
\def\confYear{2025}

\title{Enhancing Quantum-ready QUBO-based Suppression for Object Detection with Appearance and Confidence Features}

\author{
Keiichiro Yamamura$^1$\thanks{Corresponding author.} 
\and
Toru Mitsutake$^1$
\and
Hiroki Ishikura$^1$
\and
Katsuki Fujisawa$^1$
\and
Daiki Kusuhara$^2$
\and
Akihiro Yoshida$^2$
\and
$^1$ Institute of Integrated Research, Institute of Science Tokyo
\and
$^2$ Kyushu University
}

\begin{document}
\maketitle
\input{sec/abstract}
\section{Introduction}
\input{sec/introduction}
\section{Preliminaries}
\input{sec/preliminaries.tex}
\section{Method}
\input{sec/method}
\section{Experiments}
\input{sec/experiments/setting}
\input{sec/experiments/classical_computer}
\input{sec/experiments/qaqs}
\input{sec/experiments/ssim}
\section{Discussion}
\input{sec/discussion}
\section{Related Work}
\input{sec/related_work}
\section{Conclusion}
\input{sec/conclusion}

\appendix
\section*{Appendix}

\section{Example usage of our software with quantum circuits}
\label{appendix:software}
\begin{figure}[H]
    \centering
    \begin{subfigure}[b]{\linewidth}
         \centering
         \includegraphics[width=\linewidth]{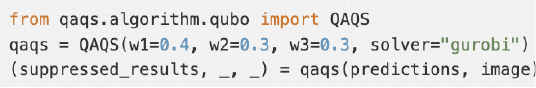}
         \caption{Run QAQS with Gurobi Optimizer.}\label{fig:gurobi}
     \end{subfigure}
    \begin{subfigure}[b]{\linewidth}
         \centering
         \includegraphics[width=\linewidth]{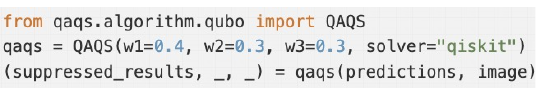}
         \caption{Run QAQS with QAOA, which depends on the quantum circuit.}\label{fig:qiskit}
     \end{subfigure}
    \caption{Example usage of our software.}
    \label{fig:example}
\end{figure}
\begin{figure}
    \centering
    \includegraphics[width=\linewidth]{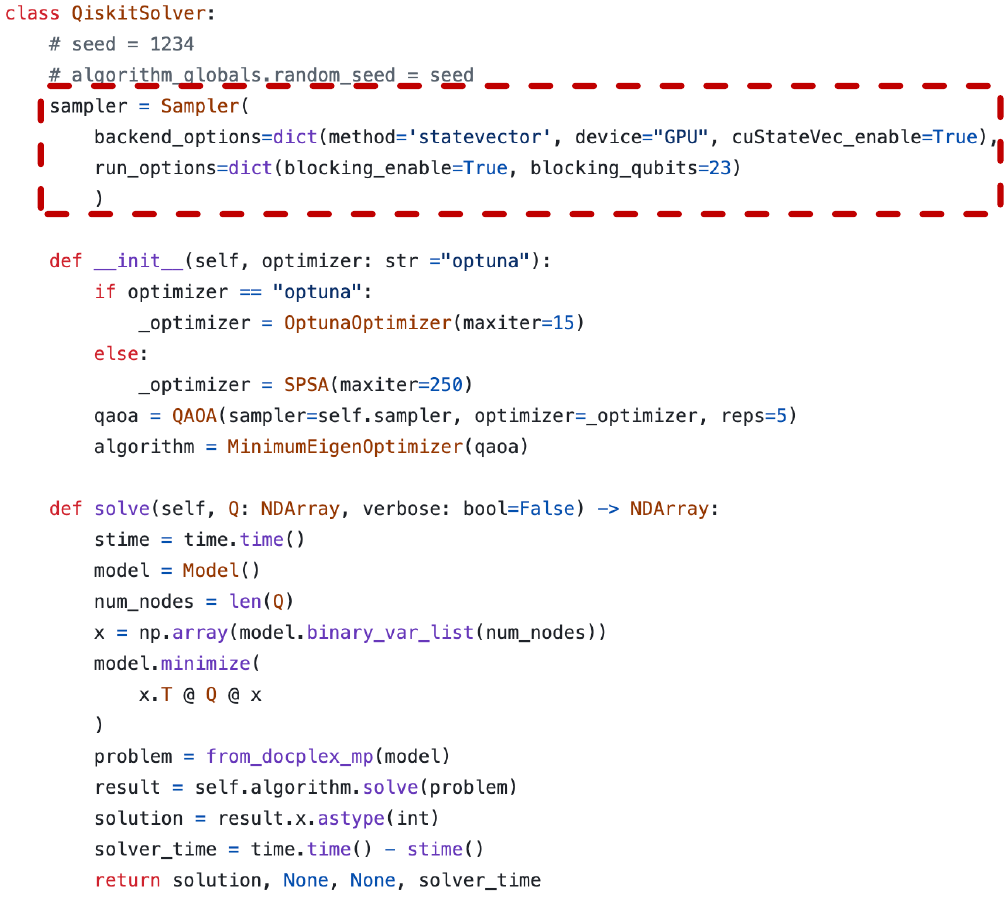}
    \caption{Implementation of QiskitSolver used to solve QUBO using QAOA algorithm.}
    \label{fig:qiskit-solver_impl}
\end{figure}

We implement our proposed methods as modularized software for portability.
A minimal sample codes are shown in \cref{fig:example}. Our implementation supports not only QAQS but also QSQS and QAQS-C.
QUBO-based suppression must be instantiated as shown in \cref{fig:example}.
The solver argument should be passed to initialize the instance. The supported solvers are the Gurobi Optimizer (\cref{fig:gurobi}) and the QAOA (\cref{fig:qiskit}). We implement QAOA using qiskit~\cite{qiskit2024}.
Qiskit enables us to switch types of quantum circuits from simulators to actual QPUs.
The class implementation for solving QUBO by the QAOA using qiskit is shown in \cref{fig:qiskit-solver_impl}. The backend of the quantum circuit is specified as the \textit{Sampler} in the area enclosed by the red dotted rectangle. This implementation uses a quantum circuit simulator running on a GPU, but a QPU published by IBM can also be specified as the backend.\footnote{https://docs.quantum.ibm.com/}

\section{Experiments including soft-scoring after solving QUBO}
\label{appendix:exp_original_impl}
In this section, we present experimental results in three different settings from the main text, shown in \cref{tab:exp_setting}.
The QSQS procedure can be decomposed into three stages: pre-processing of predictions, solving QUBO, and Soft-NMS.
\Cref{tab:exp_setting} associates the type of pre-processing and the presence or absence of Soft-NMS with the experiment number (No. column in the table).
Here, Soft-NMS refers to the processing in the blue-highlighted part of the \cref{algo:qsqs}.
For these experiments, we use the same datasets, models, and computer specifications. The hyperparameters of QUBO-based suppressions are also the same.
In experiments No. 1 and 3, we use the same parameters used by~\citet{li2020qsqs}. The IoU threshold of NMS is $0.5$, and the hyperparameters of Soft-NMS are $\sigma=0.5, O_t=0.01$.
The results of the experiments corresponding to experiments No. 1, 2, and 3 can be found in~\cref{tab:original_qsqs_v1,tab:original_qsqs_v2,tab:original_qsqs_v3}, respectively.
The main results of this experiment are summarized as follows.

\begin{itemize}
    \item Regardless of the type of preprocessing before solving QUBO, QAQS-C is robust to the presence or absence of Soft-NMS, while QSQS is significantly affected by the presence or absence of Soft-NMS. (The experiments in the main text and experiment No. 1, experiments No. 2 and 3)
     \item With the confidence score-based preprocessing before solving QUBO, Soft-NMS obscures the superiority between the different QUBO formulations. (The experiments in the main text and experiment No. 1)
    \item With NMS-based preprocessing before solving QUBO, QAQS-C tends to perform better than QSQS, regardless of the presence or absence of Soft-NMS after solving QUBO. (Experiments No. 2 and 3)
    \item With NMS-based preprocessing before solving QUBO and without Soft-NMS after QUBO, QAQS performs significantly better than QSQS. (Experiment No. 2)
    \item  With NMS-based preprocessing before solving QUBO and Soft-NMS after QUBO, QAQS is competitive with QSQS, but slightly improves mAR. (Experiment No. 3)
\end{itemize}
While the performance of QSQS relies heavily on Soft-NMS, QAQS-C can achieve high performance without Soft-NMS. This supports the superiority of our new QUBO formulations.
\begin{table}[]
    \centering
    \small
    \begin{tabular}{c ccc}
        \toprule
        No. & Before solving QUBO & After solving QUBO \\
        \midrule
        1. & Confidence score $\geq 0.25$ & Soft-NMS \\
        2. & NMS (IoU threshold $= 0.5$)  & N/A \\
        3. & NMS (IoU threshold $= 0.5$)  & Soft-NMS \\
        \bottomrule
    \end{tabular}
    \caption{Experimental settings.}
    \label{tab:exp_setting}
\end{table}
\begin{algorithm}
\centering
    \caption{QSQS-based suppression}
    \label{algo:qsqs}
    \begin{algorithmic}
        \Require{Predictions: $B\gets\{\bm{b}_1,\bm{b_2},\ldots,\bm{b}_n\}$, Confidence scores: $V\gets\{v_1,v_2,\ldots,v_n\}$, Image: $X\in\R^{C\times W\times H}$ \textcolor{blue}{Hyperparameters of soft-scoring: $O_t$ and $\sigma$}}
        \Ensure{Suppressed predictions: $D\subset B$}
        \State{Prepare $Q\in\R^{n\times n}$ using $B, V$, and $X$.}
        \State{$\displaystyle \bm{x}^* \gets\underset{\bm{x}\in\{0,1\}^{n\times 1}}{\mathrm{argmin}}\bm{x}^\top Q \bm{x}$.}\Comment{Solve QUBO}
        \State{$B_{kept}\gets\{\bm{b}_i\in B\mid \bm{x}^{*}_i = 1\}$}
        \State{$B_{soft}\gets\{\bm{b}_i\in B\mid \bm{x}^{*}_i = 0\}$}
        \State{$D\gets D\cup B_{kept}$}
        \textcolor{blue}{
        \For{$\bm{b}_i\in B_{soft}$}
            \State{$\displaystyle \bm{b}_{m}\gets\underset{\bm{b}_m\in B_{kept}}{\mathrm{argmax}}IoU(\bm{b}_i, \bm{b}_m)$}
            \State{$v_i\gets v_i \exp\left(-\frac{IoU(\bm{b}_i, \bm{b}_m)^2}{\sigma}\right)$}
            \If{$v_i\geq O_t$}
                \State{$D\gets D\cup \{\bm{b}_{i}\}$}
            \EndIf
        \EndFor
        }
    \end{algorithmic}
\end{algorithm}
\setlength\tabcolsep{0.75mm} 
\begin{table}[tbh]
    \centering
    \small
    \begin{tabular}{l ccc ccc}
    \toprule
     & \multicolumn{3}{c}{COCO 2017} & \multicolumn{3}{c}{CrowdHuman} \\
        \cmidrule(lr){1-4}
        \cmidrule(lr){5-7}
        Method    & QSQS  & QAQS  & QAQS-C & QSQS  & QAQS  & QAQS-C  \\
        \cmidrule(lr){1-4}
        \cmidrule(lr){5-7}
        mAP       & 35.36 & 35.36 & 35.36  & 35.77 & 35.77 & 35.77  \\
        mAP@50    & 55.51 & 55.51 & 55.51  & 62.44 & 62.44 & 62.44  \\
        mAP@75    & 38.24 & 38.24 & 38.24  & 36.14 & 36.14 & 36.14  \\
        mAP@S     & 14.84 & 14.84 & 14.84  & 10.08 & 10.08 & 10.08  \\
        mAP@M     & 33.99 & 33.99 & 33.99  & 30.38 & 30.38 & 30.38  \\
        mAP@L     & 46.81 & 46.81 & 46.81  & 51.81 & 51.81 & 51.81  \\
        \cmidrule(lr){1-4}
        \cmidrule(lr){5-7}
        mAR@1     & 29.26 & 29.26 & 29.26  &  3.36 &  3.36 &  3.36  \\
        mAR@10    & 44.39 & 44.39 & 44.39  & 24.46 & 24.46 & 24.46  \\
        mAR@100   & 45.92 & 45.92 & 45.92  & 42.91 & 42.91 & 42.91  \\
        mAR@S     & 23.30 & 23.30 & 23.30  & 17.78 & 17.78 & 17.78  \\
        mAR@M     & 45.18 & 45.18 & 45.18  & 39.59 & 39.59 & 39.59  \\
        mAR@L     & 56.57 & 56.57 & 56.57  & 58.55 & 58.55 & 58.55  \\
    \bottomrule
    \end{tabular}
    \caption{Results of experiment No. 1, with the confidence-based preprocessing and Soft-NMS. All the compared methods show the same performance on each dataset.}
    \label{tab:original_qsqs_v1}
\end{table}
\setlength\tabcolsep{0.75mm} 
\begin{table}[tbh]
    \centering
    \small
    \begin{tabular}{l ccc ccc}
    \toprule
     & \multicolumn{3}{c}{COCO} & \multicolumn{3}{c}{CrowdHuman} \\
        \cmidrule(lr){1-4}
        \cmidrule(lr){5-7}
        Method    & QSQS  & QAQS  & QAQS-C & QSQS  & QAQS  & QAQS-C  \\
        \cmidrule(lr){1-4}
        \cmidrule(lr){5-7}
        mAP       & 34.62 & \underline{35.97} & \textbf{36.30}  & 31.39 & \underline{36.15} & \textbf{36.87}   \\
        mAP@50    & 54.20 & \underline{56.87} & \textbf{57.51}  & 53.92 & \underline{63.69} & \textbf{65.50}   \\
        mAP@75    & 37.66 & \underline{38.77} & \textbf{38.96}  & 32.30 & \underline{36.54} & \textbf{36.81}   \\
        mAP@S     & 15.35 & \underline{15.52} & \textbf{15.63}  &  9.78 & \underline{10.62} & \textbf{10.90}   \\
        mAP@M     & 33.61 & \underline{34.59} & \textbf{34.99}  & 27.28 & \underline{30.88} & \textbf{31.58}   \\
        mAP@L     & 44.93 & \underline{47.42} & \textbf{47.84}  & 43.78 & \underline{51.62} & \textbf{52.58}   \\
        \cmidrule(lr){1-4}
        \cmidrule(lr){5-7}
        mAR@1     & 30.19 & \underline{30.61} & \textbf{30.62}  &  \underline{3.34} & \textbf{ 3.36} & \textbf{ 3.36}   \\
        mAR@10    & 44.56 & \underline{47.13} & \textbf{48.22}  & 23.36 & \underline{24.46} & \textbf{24.49}   \\
        mAR@100   & 45.44 & \underline{48.67} & \textbf{50.54}  & 36.51 & \underline{43.58} & \textbf{45.19}   \\
        mAR@S     & 26.10 & \underline{27.48} & \textbf{28.87}  & 16.72 & \underline{19.68} & \textbf{21.17}   \\
        mAR@M     & 45.31 & \underline{48.08} & \textbf{50.10}  & 34.18 & \underline{40.53} & \textbf{42.18}   \\
        mAR@L     & 54.12 & \underline{58.89} & \textbf{60.86}  & 48.57 & \underline{58.38} & \textbf{59.99}   \\
    \bottomrule
    \end{tabular}
    \caption{Results of experiment No. 2 with NMS-based preprocessing and without Soft-NMS. The best values are shown in \textbf{bold}, and the second-best values are \underline{underlined}. The proposed methods outperform QSQS.}
    \label{tab:original_qsqs_v2}
\end{table}
\setlength\tabcolsep{0.75mm} 
\begin{table}[tbh]
    \centering
    \small
    \begin{tabular}{l ccc ccc}
    \toprule
     & \multicolumn{3}{c}{COCO} & \multicolumn{3}{c}{CrowdHuman} \\
        \cmidrule(lr){1-4}
        \cmidrule(lr){5-7}
        Method    & QSQS  & QAQS  & QAQS-C & QSQS  & QAQS  & QAQS-C  \\
        \cmidrule(lr){1-4}
        \cmidrule(lr){5-7}
        mAP       & 36.25 & 36.25 & \textbf{36.30}  & 36.76 & 36.76 & \textbf{36.87}   \\
        mAP@50    & 57.41 & \underline{57.42} & \textbf{57.52}  & 65.09 & 65.09 & \textbf{65.49}   \\
        mAP@75    & 38.92 & \underline{38.93} & \textbf{38.96}  & \textbf{36.81} & \textbf{36.81} & \textbf{36.81}   \\
        mAP@S     & 15.61 & 15.61 & \textbf{15.63}  & 10.89 & 10.89 & \textbf{10.90}   \\
        mAP@M     & 34.93 & 34.93 & \textbf{34.99}  & 31.58 & 31.58 & \textbf{31.58}   \\
        mAP@L     & 47.80 & \underline{47.81} & \textbf{47.85}  & 52.52 & 52.52 & \textbf{52.58}   \\
        \cmidrule(lr){1-4}
        \cmidrule(lr){5-7}
        mAR@1     & \textbf{30.62} & \textbf{30.62} & \textbf{30.62}  &  \textbf{3.36} &  \textbf{3.36} &  \textbf{3.36}   \\
        mAR@10    & 47.97 & \underline{47.98} & \textbf{48.22}  & \textbf{24.49} & \textbf{24.49} & \textbf{24.49}   \\
        mAR@100   & 50.20 & \underline{50.23} & \textbf{50.57}  & 45.05 & \underline{45.06} & \textbf{45.19}   \\
        mAR@S     & 28.64 & \underline{28.67} & \textbf{28.88}  & 21.01 & \underline{21.04} & \textbf{21.17}   \\
        mAR@M     & 49.67 & \underline{49.68} & \textbf{50.11}  & 42.06 & \underline{42.07} & \textbf{42.18}   \\
        mAR@L     & 60.58 & \underline{60.63} & \textbf{60.90}  & 59.83 & 59.83 & \textbf{59.99}   \\
    \bottomrule
    \end{tabular}
    \caption{Results of experiment No. 3 with NMS-based preprocessing and Soft-NMS. The best values are shown in \textbf{bold}, and the second-best values are \underline{underlined}. However, the value of the second-best tie is \textbf{not underlined}. The proposed methods show slightly better performance than QSQS.}
    \label{tab:original_qsqs_v3}
\end{table}

\section{Details and experimental results for Section 5.2 in the main text}
\label{appendix:discussion}
\begin{table*}[tbh]
    \centering
    \small
    \begin{tabular}{l cc c cc c cc c  cc c cc c cc}
    \toprule
     & \multicolumn{8}{c}{COCO} && \multicolumn{8}{c}{CrowdHuman} \\
        \cmidrule(lr){1-1}
        \cmidrule(lr){2-9}
        \cmidrule(lr){10-18}
        Method    & \multicolumn{2}{c}{QSQS} && \multicolumn{2}{c}{QAQS} && \multicolumn{2}{c}{QAQS-C} && \multicolumn{2}{c}{QSQS} && \multicolumn{2}{c}{QAQS} && \multicolumn{2}{c}{QAQS-C} \\
        \cmidrule(lr){1-1}
        \cmidrule(lr){2-3}
        \cmidrule(lr){5-6}
        \cmidrule(lr){8-9}
        \cmidrule(lr){11-12}
        \cmidrule(lr){14-15}
        \cmidrule(lr){17-18}
        Time [s]  & 108   & 59    && 122   & 63    && 121   & 58     && 140   &    99 && 163   &   100 && 155   & 95    \\
        Enforce sparsity$^*$  & -   & $\checkmark$    &&  -   & $\checkmark$   && -   & $\checkmark$      &&  -   & $\checkmark$  &&  -   & $\checkmark$  &&  -   & $\checkmark$    \\
        \cmidrule(lr){1-1}
        \cmidrule(lr){2-3}
        \cmidrule(lr){5-6}
        \cmidrule(lr){8-9}
        \cmidrule(lr){11-12}
        \cmidrule(lr){14-15}
        \cmidrule(lr){17-18}
        mAP       & 34.04 & 34.02 && 35.25 & 35.27 && 35.35 & 35.36  && 31.23 & 32.92 && 35.62 & 35.73 && 35.77 & 35.77 \\
        mAP@50    & 52.94 & 53.07 && 55.33 & 55.37 && 55.50 & 55.50  && 53.28 & 56.71 && 61.89 & 62.49 && 62.44 & 62.44 \\
        mAP@75    & 37.19 & 37.15 && 38.17 & 38.18 && 38.22 & 38.23  && 32.32 & 34.34 && 36.15 & 36.15 && 36.14 & 36.14 \\
        mAP@S     & 14.63 & 14.65 && 14.79 & 14.79 && 14.83 & 14.83  &&  9.42 &  9.63 && 10.09 & 10.08 && 10.08 & 10.08 \\
        mAP@M     & 32.91 & 33.07 && 33.88 & 33.91 && 34.02 & 34.02  && 27.11 & 28.41 && 30.35 & 30.41 && 30.38 & 30.38 \\
        mAP@L     & 44.43 & 44.18 && 46.64 & 46.65 && 46.78 & 46.79  && 43.79 & 46.63 && 51.55 & 51.54 && 51.81 & 51.81 \\
        \cmidrule(lr){1-1}
        \cmidrule(lr){2-3}
        \cmidrule(lr){5-6}
        \cmidrule(lr){8-9}
        \cmidrule(lr){11-12}
        \cmidrule(lr){14-15}
        \cmidrule(lr){17-18}        
        mAR@1     & 28.92 & 28.79 && 29.26 & 29.26 && 29.26 & 29.26  &&  3.34 &  3.35 &&  3.36 &  3.36 &&  3.36 &  3.36 \\
        mAR@10    & 42.13 & 42.14 && 44.19 & 44.21 && 44.38 & 44.39  && 23.38 & 23.73 && 24.45 & 24.45 && 24.46 & 24.46 \\
        mAR@100   & 42.90 & 43.04 && 45.50 & 45.57 && 45.90 & 45.91  && 36.19 & 38.45 && 42.57 & 42.67 && 42.91 & 42.91 \\
        mAR@S     & 22.10 & 22.22 && 23.01 & 23.05 && 23.30 & 23.30  && 15.55 & 16.08 && 17.56 & 17.61 && 17.78 & 17.78 \\
        mAR@M     & 42.47 & 42.81 && 44.76 & 44.85 && 45.17 & 45.17  && 33.86 & 35.81 && 39.29 & 39.38 && 39.59 & 39.59 \\
        mAR@L     & 52.14 & 52.00 && 56.07 & 56.11 && 56.52 & 56.53  && 48.66 & 52.08 && 58.13 & 58.26 && 58.55 & 58.55 \\
    \bottomrule
    \end{tabular}
    \\\footnotesize{$*$ Sparse coefficient matrix using $P_1$ and $P_2$ defined in \cref{eq:sparse_p1,eq:sparse_p2}.}
    \caption{Comparison with default coefficient matrix and sparse coefficient matrix.}
    \label{tab:default_vs_sparse}
\end{table*}
In this section, we will discuss in more detail the acceleration based on improving the sparsity of the coefficient matrix, which is mentioned in Section 5.2 of the main text.
First, we explain the method of calculating the coefficient matrix to improve its sparsity.
Generalized IoU (GIoU) is defined as follows.
\begin{equation}
    GIoU(A, B) = IoU(A, B) - \frac{|C(A, B)\setminus A\cup B|}{|C(A, B)|},
\end{equation}
where $C(A, B)$ is the minimum convex hull that covers $A\cup B$.
We use this GIoU instead of IoU to improve the sparsity of the coefficient matrix.
Although IoU is greater than or equal to 0 for any inputs, GIoU can take negative values.
To adjust the range of coefficient to the original IoU version, we clip the negative value of GIoU to 0.
From the definition of GIoU, the process of clipping at $0$ if $GIoU(A, B) \leq 0$ is equivalent to ignoring prediction overlap when IoU is smaller than the threshold of $\dfrac{|C\setminus A\cup B|}{|C|}$.
Also, in the original formulation using IoU, the components of the spatial feature $(P_2)_{ij}$ is 0 when $(P_1)_{ij}=0$, i.e., IoU equals 0.
To maintain consistency with this, we decide to replace the corresponding components of $P_2$ with 0 for pairs that make GIoU equal to 0.
To summarize the above, the sparse version of $P_1$ and $P_2$ are defined as follows.
\begin{align}
    (P_1)_{ij} &= \left\{
        \begin{array}{cl}
            IoU(b_i, b_j) & \mathrm{if}~IoU(b_i, b_j) \geq \dfrac{|C(b_i, b_j)\setminus b_i\cup b_j|}{|C(b_i, b_j)|}\\
            0 & \mathrm{otherwise}
        \end{array}\right.\label{eq:sparse_p1}\\
    (P_2)_{ij} &= \left\{
        \begin{array}{cl}
            \dfrac{|b_i\cap b_j|}{\sqrt{|b_i||b_j|}} & \mathrm{if}~IoU(b_i, b_j) \geq \dfrac{|C(b_i, b_j)\setminus b_i\cup b_j|}{|C(b_i, b_j)|}\\
            0 & \mathrm{otherwise}
        \end{array}\right.\label{eq:sparse_p2}
\end{align}
The modified coefficient matrix has the same or fewer non-zero components than the original because $IoU(A, B)\leq GIoU(A, B)$ when $C\ne A\cup B$. This means the modified coefficient matrix is more sparse.
The experimental results are shown in \cref{tab:default_vs_sparse}.
The experimental settings are the same as in the main text.
Regardless of the dataset, the higher the sparsity of the coefficient matrix, the shorter the computation time, with up to around 50\% reduction of the execution time.
In most cases, we observe improvements in computation speed while maintaining performance or with a slight improvement.
One exception is QSQS with the COCO dataset. QSQS improves the computation speed at the expense of a slight decrease in performance.

{
    \small
    \bibliographystyle{ieeenat_fullname}
    \bibliography{reference}
}


\end{document}

%% file: math_commands.tex
\usepackage{amsmath,amsfonts,amssymb,bm}

\def\eqref#1{equation~\ref{#1}}

\def\1{\bm{1}}

\DeclareMathAlphabet{\mathsfit}{\encodingdefault}{\sfdefault}{m}{sl}
\SetMathAlphabet{\mathsfit}{bold}{\encodingdefault}{\sfdefault}{bx}{n}

\newcommand{\R}{\mathbb{R}}



%% file: sec/abstract.tex
\begin{abstract}
    Quadratic Unconstrained Binary Optimization (QUBO)-based suppression in object detection is known to have superiority to conventional Non-Maximum Suppression (NMS), especially for crowded scenes where NMS possibly suppresses the (partially-) occluded true positives with low confidence scores.
    Whereas existing QUBO formulations are less likely to miss occluded objects than NMS, there is room for improvement because existing QUBO formulations naively consider confidence scores and pairwise scores based on spatial overlap between predictions.
    This study proposes new QUBO formulations that aim to distinguish whether the overlap between predictions is due to the occlusion of objects or due to redundancy in prediction, i.e., multiple predictions for a single object.
    The proposed QUBO formulation integrates two features into the pairwise score of the existing QUBO formulation: i) the appearance feature calculated by the image similarity metric and ii) the product of confidence scores.
    These features are derived from the hypothesis that redundant predictions share a similar appearance feature and (partially-) occluded objects have low confidence scores, respectively.
    The proposed methods demonstrate significant advancement over state-of-the-art QUBO-based suppression without a notable increase in runtime, achieving up to 4.54 points improvement in mAP and 9.89 points gain in mAR.
\end{abstract}

%% file: sec/introduction.tex
Quantum computers, which are expected to have quantum supremacy~\cite{Arute2019} over classical computers, have attracted significant attention, and the number of physical qubits has been steadily increasing.
Although the number of logical qubits in publicly available quantum computers is currently limited, exploring its potential for industrial applications in advance is extremely important.
One application of quantum computers is the suppression of object detection using deep learning models.
Object detection, which predicts the category and location of objects in images, has various industrial applications, including facial recognition~\cite{YangLLT16} and pedestrian detection for surveillance~\cite{Dominguez-Sanchez18}, automatic driving~\cite{GeigerLU12}, and robots~\cite{MeiHPT15}.

Deep learning-based object detectors predict the category and location of objects from image features extracted by neural networks such as convolutional neural networks~\cite{SimonyanZ14aVGG,HeZRS16resnet} and vision transformers~\cite{DosovitskiyB0WZ21ViT,LiuL00W0LG21Swin}.
Most detectors are trained to detect objects from each predefined region (anchor-based) or each point in the feature map (anchor-free).
Therefore, the detectors likely output redundant detections, i.e., multiple predictions for a single object. Non-Maximum Suppression (NMS) has been used to suppress such redundant detections for a long time.

However, NMS-based greedy suppression is known to miss the partially occluded objects.
To address this issue, NMS variants~\cite{bodla2017snms,He2019SofterNMS,Liu2019AdaNMS,Nils2020vgnms,Huang2020VFGNMS,shepley2023confluence}, clustering-based suppression~\cite{ShenJXLK22CPCluster}, and Quadratic Unconstrained Binary Optimization (QUBO)-based suppression~\cite{rujikietgumjorn2013qubo,li2020qsqs} have been studied.
QUBO is a type of combinatorial optimization problem where quantum optimization algorithms are expected to find high-quality approximate solutions quickly.
We focus on refining the QUBO-based suppression to develop a quantum-ready method that can be integrated with a quantum computer.

QUBO-based suppression was originally proposed as an effective suppression for pedestrian detection where the objects are likely to be occluded~\cite{rujikietgumjorn2013qubo}.
\citet{li2020qsqs} proposed Quantum-soft QUBO Suppression (QSQS), a hybrid classical-quantum algorithm that combines QUBO-based suppression with soft-scoring similar to Soft-NMS~\cite{bodla2017snms}. 
QSQS is the state-of-the-art (SOTA) QUBO-based suppression that has demonstrated superiority over NMS not only in crowded scenes but also in general non-crowded scenes.
Despite this great advancement, there is room for improvement because the QUBO formulation of QSQS considers only the confidence score and pairwise score based on the spatial overlap between predictions.

In this study, we propose enhanced QUBO formulations that distinguish predictions for partially occluded objects from purely redundant predictions.
The proposed QUBO formulations integrate two features into the QUBO formulation of QSQS: i) the appearance feature that reflects image similarity and ii) the product of confidence scores between prediction pairs.
This integration is based on the hypothesis that the visual appearance and confidence score reflect the difference between predictions for partially occluded objects and redundant predictions.
We use Structural SIMilarity (SSIM)~\cite{wang2004ssim} as the appearance feature because SSIM has beneficial characteristics to construct the QUBO coefficient matrix.
To reduce the computation cost of SSIM, we utilize the divide-and-conquer algorithm~\cite{doi:10.1137/S0895479892241287,Dwyer1987} and GPU parallelization. The proposed SSIM implementation reduces the runtime from 2 and 11~s/image (seconds per image) to 6 and 14~ms/image (milliseconds per image) for non-crowded and crowded scenes, respectively, under approximately 1~GB GPU memory usage.

The proposed methods consistently outperform QSQS, the SOTA QUBO-based suppression, on CrowdHuman~\cite{shao2018crowdhuman} dataset for crowded and on COCO~\cite{lin2015microsoft} dataset for non-crowded scenes.
The proposed methods especially show superior performance on the crowded dataset.
To fairly compare the advantages among QUBO formulations, we directly compare the results only with the QUBO solution, not including the results of soft-scoring. 
Our experiments use Gurobi Optimizer (Gurobi)~\cite{gurobi}, a branch-and-bound-based classical solver, instead of quantum solvers because the currently available quantum gate computers have only a few valid logical qubits.
However, our software is designed to integrate quantum computers, i.e., quantum-ready. See Supplementary A for usage of our software.
We summarize our contributions below.
\begin{enumerate}
    \item We propose new formulations for QUBO-based suppression that integrate the appearance feature and product of confidence scores between predictions.
    \item We propose a faster implementation of SSIM, which is used as the appearance feature, and reduce the runtime from up to 11~s/image to up to 14~ms/image.
    \item We publish a quantum-ready software that can be integrated with quantum computers after paper acceptance.
    \item The proposed formulations outperform QSQS, the SOTA QUBO-based suppression, achieving up to 4.54 points improvement in mAP and 9.89 points gain in mAR without a notable increase in runtime.
\end{enumerate}

%% file: sec/preliminaries.tex
\subsection{Quadratic Unconstrained Binary Optimization}
Quadratic Unconstrained Binary Optimization (QUBO) is a combinatorial optimization problem formulated as follows.
\begin{equation}
    \max_{\bm{x}\in\{0, 1\}^n}\bm{x}^\top Q\bm{x} = \sum_{i=1}^{n}\sum_{j=1}^{n}Q_{ij}\bm{x}_{i}\bm{x}_{j}+\sum_{i=1}^{n} Q_{ii}\bm{x}_i,
\end{equation}
where $Q\in\R^{n\times n}$.
This equation arise from the fact that $\bm{x}_{i}^2=\bm{x}_{i}$ holds when $\bm{x}_i\in\{0, 1\}$.
QUBO is equivalent to the Ising model, which is efficiently optimized via quantum algorithms such as quantum annealing~\cite{PhysRevE.58.5355Nishimori,morita2008mathematical} and Quantum Approximate Optimization Algorithm (QAOA)~\cite{farhi2014quantum}.
QUBO was originally used for the suppression in pedestrian detection~\cite{rujikietgumjorn2013qubo} and extended to generic object detection later~\cite{li2020qsqs}. These studies demonstrated the superiority of QUBO-based suppression over NMS for crowded and non-crowded datasets.
We aim to propose enhanced QUBO formulations that distinguish whether the overlapped prediction is due to occluded objects or redundant prediction.
Our formulations significantly improve QUBO-based suppression accuracy for crowded and non-crowded datasets.

\vskip.5\baselineskip\noindent\textbf{QUBO framework.} 
QUBO Framework (QF)~\cite{rujikietgumjorn2013qubo} first formulate the suppression of pedestrian detection as QUBO. QF aims to reduce false positives in pedestrian detection. Although QF performed better than NMS for pedestrian detection, where occlusion is likely to occur, the authors concluded that QF is not better than NMS for non-crowded scenes. The coefficient matrix of QF is defined as follows.
\begin{equation}
    Q = w_1 L - w_2 P_1,
\end{equation}
where $L=\mathrm{diag}(\bm{v})$ is a diagonal matrix whose elements are confidence scores, $P_1$ is a symmetric matrix of pairwise scores such as Intersection over Union (IoU), and $w_1\geq0, w_2\geq0$ are hyperparameters which satisfy $w_1+w_2=1$.

\vskip.5\baselineskip\noindent\textbf{Quantum-Soft QUBO Suppression.} 
Quantum-Soft QUBO Suppression (QSQS)~\cite{li2020qsqs} is a quantum-classical hybrid algorithm that consists of QUBO with a modified coefficient matrix considering IoU and spatial feature~\cite{lee2016individual} and soft-scoring similar to Soft-NMS~\cite{bodla2017snms}.
QSQS performed better than conventional NMS for generic object detection such as PASCAL VOC 2007~\cite{Everingham15pascal} and COCO in addition to pedestrian detection.
QSQS solves QUBO first, then executes soft-scoring for predictions suppressed through QUBO to retain predictions whose score is higher than the pre-defined threshold.
The coefficient matrix for QSQS is defined as follows.
\begin{equation}
    Q = w_1L - (w_2 P_1 + w_3 P_2),
\end{equation}
where $L=\mathrm{diag}(\bm{v})$ is a diagonal matrix whose elements are confidence scores, $P_1$ and $P_2$ are symmetric matrices whose elements are IoU and spatial feature~\cite{lee2016individual,li2020qsqs}, and $w_1\geq0, w_2\geq0, w_3\geq0$ are hyperparameters which satisfy $w_1+w_2+w_3=1$.

\subsection{Structural similarity}
Structural SIMilarity (SSIM)~\cite{wang2004ssim,hore2010psnr_ssim} is based on the similarity of the brightness, contrast, and structure of image patches and is used for image quality evaluation.
SSIM between corresponding image patches ($\bm{\mathrm{x}}$ and $\bm{\mathrm{y}}$) of two images ($X$ and $Y$) is defined as follows.
\begin{align}
    & SSIM(\bm{\mathrm{x}}, \bm{\mathrm{y}}) \\
    & = l(\bm{\mathrm{x}}, \bm{\mathrm{y}})\cdot c(\bm{\mathrm{x}}, \bm{\mathrm{y}}) \cdot s(\bm{\mathrm{x}}, \bm{\mathrm{y}}) \\
    & = \left(\dfrac{2\mu_{\bm{\mathrm{x}}}\mu_{\bm{\mathrm{y}}}+C_1}{\mu_{\bm{\mathrm{x}}}^2+\mu_{\bm{\mathrm{y}}}^2+C_1}\right)\cdot \left(\dfrac{2\sigma_{\bm{\mathrm{x}}}\sigma_{\bm{\mathrm{y}}} + C_2}{\sigma_{\bm{\mathrm{x}}}^2+\sigma_{\bm{\mathrm{y}}}^2+C_2}\right) \cdot \left(\dfrac{2\sigma_{\bm{\mathrm{x}}\bm{\mathrm{y}}}+C_3}{\sigma_{\bm{\mathrm{x}}}\sigma_{\bm{\mathrm{y}}}+C_3}\right),
\end{align}
where $C_1=0.01^2, C_2=0.03^2, C_3=2C_2$ are parameters to mitigate tiny denominator, and $\mu_{\bm{\mathrm{x}}}$ and $\sigma_{\bm{\mathrm{x}}}$ represents the mean and standard deviation over image patch $\bm{\mathrm{x}}$, respectively.
SSIM between two images $X$ and $Y$ is defined as the average of SSIM for image patches over a set of image patches $W$, i.e., $\frac{1}{|W|}\sum_{w\in W}SSIM(\bm{\mathrm{x}}_w, \bm{\mathrm{y}}_w)$. We call this value as SSIM.
SSIM has the following characteristics.
\begin{enumerate}
    \item Symmetric: $SSIM(\bm{\mathrm{x}}, \bm{\mathrm{y}})=SSIM(\bm{\mathrm{y}}, \bm{\mathrm{x}})$.
    \item Upper bounded: $SSIM(\bm{\mathrm{x}}, \bm{\mathrm{y}})\leq 1$.
    \item $SSIM(\bm{\mathrm{x}}, \bm{\mathrm{y}})=1$ if and only if $\bm{\mathrm{x}} = \bm{\mathrm{y}}$.
\end{enumerate}

%% file: sec/method.tex
The QUBO-based suppression was originally developed for pedestrian detection, where objects are likely to overlap~\cite{rujikietgumjorn2013qubo}.
Although QUBO-based suppression worked better than NMS, they may suppress the occluded true positives because they naively utilize the pairwise score related to the overlap between predictions. 
We devise new coefficient matrices based on the following two hypotheses to distinguish the occluded true positives and redundant predictions.
\begin{enumerate}
    \item Redundant predictions are distinguished from occluded true positives by considering the appearance feature of predictions.
    \item Confidence scores should be integrated into pairwise scores because occluded objects may have lower confidence scores.
\end{enumerate}
Furthermore, we propose a faster implementation of SSIM, which is used as the appearance feature, to improve the overall throughput of object detection.

\input{sec/method_QAQS}
\input{sec/method_SSIM}

%% file: sec/method_QAQS.tex

\subsection{Construction of coefficient matrix}
\noindent\textbf{Quantum Appearance QUBO Suppression (QAQS)} \\
We introduce an appearance feature to the non-diagonal elements of the QSQS coefficient matrix to distinguish redundant prediction from occluded objects. This modification is based on the hypothesis that redundant predictions share a similar image feature. The coefficient matrix $Q$ is defined as follows.
\begin{equation}
    Q = w_1L - (w_2 P_1 + w_3 P_2)\odot A,\label{eq:qaqs}
\end{equation}
where $L=\mathrm{diag}(\bm{v})$ is a diagonal matrix whose elements are confidence scores, $P_1$ and $P_2$ are the same as in QSQS, symmetric matrices whose elements are IoU and spatial feature, $A$ is a symmetric matrix of appearance feature, and $w_1\geq0, w_2\geq0, w_3\geq0$ are hyperparameters which satisfy $w_1+w_2+w_3=1$. $\odot$ denotes the element-wise product.

We choose SSIM as the appearance feature for the following reasons. First, the SSIM value is less than or equal to one for arbitrary image pairs. This feature is desirable to maintain the scale of $Q$ within a reasonable range. Second, SSIM is symmetric and equal to one only if the two images are identical. IoU and the spatial feature have the same characteristics. Finally, SSIM is efficiently computed on GPUs, as detailed in \cref{sec:ssim_impl}.

\vskip.5\baselineskip\noindent\textbf{QAQS with Confidence-based weighting (QAQS-C)} \\
In addition to introducing an appearance feature, we further incorporate the confidence score into the non-diagonal part of the coefficient matrix based on the hypothesis that (partially-) occluded objects have low confidence scores.
DPP~\cite{lee2016individual}, which integrates the confidence score to the similarity term, also motivates this modification. The final formulation is provided as follows.
\begin{equation}
    Q = w_1L - \bm{v}\left((w_2 P_1 + w_3 P_2)\odot A\right)\bm{v}^\top.\label{eq:qaqs_c}
\end{equation}
Notations are the same as QAQS. Here, $\bm{v}\in\R^{n \times 1}$.


%% file: sec/method_SSIM.tex
\subsection{A faster implementation of SSIM}
\label{sec:ssim_impl}
Although SSIM is usually computed for a pair of images, we have to compute SSIM for combinations of $n$ predictions, i.e., $n(n-1)/2$ pairs of predictions.
Existing software is sufficiently fast for usual purposes, but we need a much faster implementation. 
To achieve this, we introduce the three techniques: parallelization on GPU, divide-and-conquer algorithm to reduce memory usage, and omission of redundant computation. 

\vskip.5\baselineskip\noindent\textbf{Parallelization on GPU. }
SSIM computation is easily parallelized on GPU using the convolution operator with a two-dimensional Gaussian filter\footnote{\url{https://github.com/kornia/kornia/tree/main}}\footnote{\url{https://github.com/pytorch/ignite/blob/master/ignite/metrics/ssim.py}}.
We simultaneously compute SSIM for $n(n-1)/2$ combinations through batch parallelization by resizing all predictions to the same shape.
Empirically, resizing to around $48\times48$ does not affect performance.
This parallelization significantly reduces the computation time compared to the sequential computation of SSIM.
However, memory usage tends to be large because all data should be on GPU memory at the same time. 

\begin{figure}
    \centering
    \includegraphics[width=\linewidth]{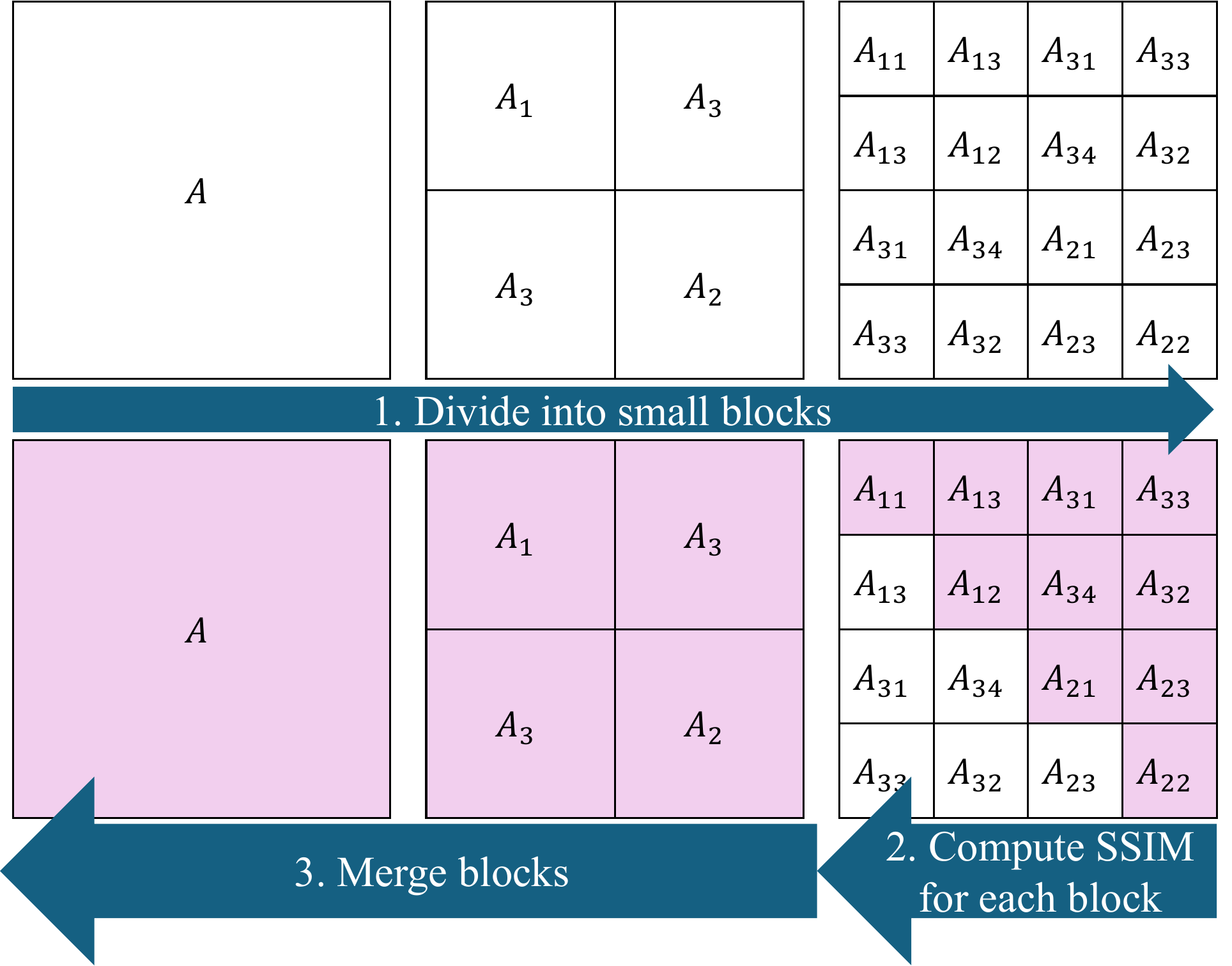}
    \caption{Illustration of our divide-and-conquer algorithm for SSIM calculation. 
    First, the matrix of appearance feature ($A$) is recursively divided into four blocks at each recursion step. Second, the SSIM values of each block are calculated in parallel when the size of the divided blocks becomes sufficiently small. Blocks of lower triangular parts are excluded from the computation target because this part can be completed using the symmetry of $A$. Finally, computation results for small blocks are merged. 
    }
    \label{fig:divideandconquer}
\end{figure}

\vskip.5\baselineskip\noindent\textbf{Divide-and-conquer algorithm to reduce memory usage. }
We use the divide-and-conquer algorithm to reduce the peak memory usage during parallelization on GPU.
Divide-and-conquer is a paradigm that solves a problem by breaking it into smaller sub-problems and solving all the sub-problems.
\Cref{fig:divideandconquer} illustrates the divide-and-conquer algorithm for our SSIM computation.
SSIM is calculated as a matrix with $n^2$ elements at once with simple GPU parallelization. However, our divide-and-conquer-based implementation first computes SSIM for sub-matrices of the overall matrix of appearance feature ($A$) and merges them later.
As a result, our implementation computes the SSIM matrix as nearly upper triangular and significantly reduces the peak GPU memory usage. 

\begin{figure}
    \centering
    \begin{subfigure}[b]{0.45\linewidth}
         \centering
         \includegraphics[width=\linewidth]{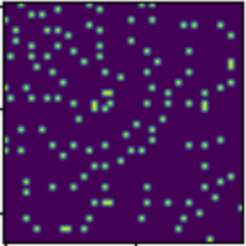}
         \caption{Before RCM algorithm.}
     \end{subfigure}
     \hfill
    \begin{subfigure}[b]{0.45\linewidth}
         \centering
         \includegraphics[width=\linewidth]{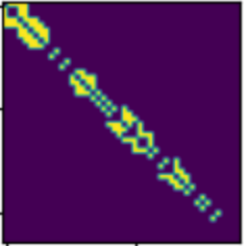}
         \caption{After RCM algorithm.}
     \end{subfigure}
    \caption{Visualization of \textit{Intersection Matrix} $I\in\{0,1\}^{n\times n}$. 
    Zero-value elements are colored in purple, and one-value elements are colored in yellow.
    }
    \label{fig:plot_intersection_mat}
\end{figure}

\vskip.5\baselineskip\noindent\textbf{Omission of redundant computation. }
The SSIM calculation for predictions with no overlap can be omitted without affecting the coefficient matrix because the corresponding pairwise scores ($P_1$ and $P_2$) are 0.
To enhance the effectiveness of this technique, we permute the order of predictions before SSIM computation.
We define the \textit{Intersection Matrix} $I\in\{0, 1\}^{n\times n}$ as a symmetric matrix whose $ij$ element takes 1 if predictions $i$ and $j$ are overlapped and otherwise 0.
Then, we permute $I$ using the Reverse Cuthill Mckee (RCM) algorithm~\cite{Cuthill1969RCM} so that the non-zero elements of $I$ are as close as possible to diagonals.
As shown in \cref{fig:plot_intersection_mat}, the permuted $I$ has large zero matrix blocks, where corresponding SSIM computations are efficiently omitted.
The RCM algorithm is based on breadth-first search and minimizes the bandwidth of the matrix, i.e., the maximum distance between non-zero and diagonal elements.
The time complexity of the RCM algorithm is $O(n+m)$, where $n$ is the number of columns and rows of $I$, and $m$ is the number of non-zero elements of $I$.
The overhead of the RCM algorithm is negligible because the number of predictions, $n$, is not so large.

%% file: sec/experiments/setting.tex
We demonstrate the superiority of our QUBO formulations for suppression over the QSQS formulation through experiments on both crowded and non-crowded datasets.
\subsection{Setting}
\noindent\textbf{Dataset and Model. }
We use the COCO 2017~\cite{lin2015microsoft} validation set (5,000 images) for non-crowded scenes and the CrowdHuman~\cite{shao2018crowdhuman} validation set (4,370 images) for crowded scenes.
Following the experimental setting in the existing study~\cite{li2020qsqs}, we implement QUBO-based suppressions on two-stage Faster R-CNN~\cite{ren2015faster_r_cnn} baseline detector with ResNet~\cite{HeZRS16resnet} backbone. While \citet{li2020qsqs} trained their model starting from the public model, we utilize pre-trained models with ResNet-50 backbone and Feature Pyramid Network (FPN)~\cite{LinDGHHB17FPN} to enhance reproducibility.
The pre-trained weight for the COCO dataset is available at PyTorch website\footnote{\url{https://pytorch.org/vision/stable/models/faster_rcnn.html}}. For the CrowdHuman dataset, we use the pre-trained weight\footnote{\url{https://github.com/aibeedetect/bfjdet}} trained by \citet{Wan2021Body}.

\vskip.5\baselineskip\noindent\textbf{Evaluation metrics. }
Similar to the existing studies~\cite{ren2015faster_r_cnn,li2020qsqs,shepley2023confluence}, we report the mean average precision (mAP) and mean average recall (mAR), a standard metric for generic object detection. The precise definition of mAP and mAR is the same as those of COCO~\cite{lin2015microsoft}. We report results for a single run because suppression methods are deterministic.

\vskip.5\baselineskip\noindent\textbf{Computer specification. }
The experiments are mainly done with Intel(R) Xeon(R) Gold 6240R CPU @ 2.40GHz and NVIDIA GeForce RTX 3090. Due to the GPU memory usage, ablation studies of SSIM computation with the CrowdHuman dataset are conducted on Intel(R) Xeon(R) Gold 5220R CPU @ 2.20GHz and NVIDIA RTX A6000.

\subsection{Implementation Details}

\vskip.5\baselineskip\noindent\textbf{Baselines. }
Our goal is to propose a better QUBO formulation to suppress redundant predictions. For this purpose, QSQS, the SOTA QUBO-based suppression, is a good baseline. However, the soft-scoring in QSQS obscures the contribution of QUBO formulation quality. Therefore, we directly compare the solutions for each QUBO formulation in our experiments.
See Supplementary B for results with soft-scoring.
For reference, we also report the experimental results with the conventional suppression algorithms, including NMS and Soft-NMS.

\vskip.5\baselineskip\noindent\textbf{Preprocessing before suppression. }
\citet{li2020qsqs} applied NMS-based preprocessing to detections before solving QUBO to prevent solving too large QUBO problems in their experiments.
Instead, we apply confidence-based preprocessing that excludes low-quality predictions whose scores are lower than 0.25 to focus on evaluating the quality of QUBO formulations. The score threshold is determined by considering the balance between quantitative and qualitative results.

\vskip.5\baselineskip\noindent\textbf{Hyperparameter. }
The hyperparameters of QUBO-based suppression, $w_1, w_2, w_3$ are tuned via a black-box optimization based on Tree-structured Parzen Estimator~\cite{BergstraBBK11TPE,BergstraYC13TPE} implemented on Optuna~\cite{AkibaSYOK19Optuna}. After a group of hyperparameter optimization, we fix these parameters to the same values as the original QSQS implementation, i.e., $w_1=0.4, w_2=0.3, w_3=0.3$.
Considering the quantitative and qualitative performance, the IoU threshold of NMS and $\sigma$ of Gaussian weighting in Soft-NMS are fixed at $0.3$ and $0.5$, respectively.

\vskip.5\baselineskip\noindent\textbf{Solver. }
The QUBO problems are solved by the exact solver, Gurobi Optimizer (Gurobi)~\cite{gurobi} version 11.0.3. Gurobi uses two CPU threads and terminates when the execution time reaches 60 seconds. In the experiments, Gurobi finds the optimal solution for each QUBO problem within 60 seconds.

%% file: sec/experiments/classical_computer.tex
\subsection{Results}

\noindent\textbf{COCO 2017 dataset. }
The left part of \cref{tab:results_faster_rcnn} shows the results on the COCO dataset. The proposed methods, QAQS and QAQS-C, outperform QSQS by 1.21 and 1.31 points in mAP and 3.93 and 4.38 points in mAR, respectively.
The proposed methods significantly improve mAP and mAR for large objects (mAP@L and mAR@L).
The higher value of mAP and mAR indicates a lower number of false positives and false negatives. These results demonstrate our approach's effectiveness in reducing the number of redundant predictions while detecting more (occluded) true positives. Our methods also show promising results compared to conventional NMS and Soft-NMS for reference.
\setlength\tabcolsep{0.75mm} 
\begin{table*}
    \centering
    \begin{tabular}{l r ccccc c ccccc}
    \toprule
     & & \multicolumn{5}{c}{COCO 2017}& & \multicolumn{5}{c}{CrowdHuman} \\
    \cmidrule(lr){3-7}
    \cmidrule(lr){9-13}
    \multicolumn{2}{c}{Metric} & NMS & Soft-NMS & QSQS$^{*}$ & QAQS$^{*,\dag}$ & QAQS-C$^{*,\dag}$ && NMS & Soft-NMS & QSQS$^{*}$ & QAQS$^{*,\dag}$ & QAQS-C$^{*,\dag}$ \\
    \cmidrule(lr){1-2}
    \cmidrule(lr){3-7}
    \cmidrule(lr){9-13}
    mAP        & ($\uparrow$) & 34.61 & 35.20 & 34.04 & \underline{35.25} & \textbf{35.35} && 34.01 & 34.97 & 31.23 & \underline{35.62} & \textbf{35.77} \\
    mAP@50     & ($\uparrow$) & 54.08 & 55.27 & 52.94 & \underline{55.33} & \textbf{55.50} && 58.47 & 61.23 & 53.28 & \underline{61.89} & \textbf{62.44} \\
    mAP@75     & ($\uparrow$) & 37.69 & 38.14 & 37.19 & \underline{38.17} & \textbf{38.22}& & 35.09 & 35.68 & 32.32 & \textbf{36.15} & \underline{36.14} \\
    mAP@S  & ($\uparrow$)     & 14.70 & \underline{14.82} & 14.63 & 14.79 & \textbf{14.83} &&  9.72 &  9.94 &  9.42 & \textbf{10.09} & \underline{10.08} \\
    mAP@M & ($\uparrow$)      & 33.22 & 33.78 & 32.91 & \underline{33.88} & \textbf{34.02} && 29.16 & 29.93 & 27.11 & \underline{30.35} & \textbf{30.38} \\
    mAP@L  & ($\uparrow$)     & 45.68 & 46.61 & 44.43 & \underline{46.64} & \textbf{46.78} && 48.50 & 50.62 & 43.79 & \underline{51.55} & \textbf{51.81} \\
    \cmidrule(lr){1-2}
    \cmidrule(lr){3-7}
    \cmidrule(lr){9-13}
    mAR@1      & ($\uparrow$) & \textbf{29.26} & \textbf{29.26} & \underline{28.92} & \textbf{29.26} & \textbf{29.26} &  &\textbf{3.36} &  \textbf{3.36} &  \underline{3.34} &  \textbf{3.36} & \textbf{ 3.36} \\
    mAR@10     & ($\uparrow$) & 42.73 & 43.92 & 42.13 & \underline{44.19} & \textbf{44.38} && 24.01 & 24.09 & 23.38 & \underline{24.45} & \textbf{24.46} \\
    mAR@100    & ($\uparrow$) & 43.79 & 45.22 & 42.90 & \underline{45.50} & \textbf{45.90} && 39.77 & 41.92 & 36.19 & \underline{42.57} & \textbf{42.91} \\
    mAR@S  & ($\uparrow$)     & 22.37 & 22.91 & 22.10 & \underline{23.01} & \textbf{23.30} && 16.41 & 17.10 & 15.55 & \underline{17.56} & \textbf{17.78} \\
    mAR@M & ($\uparrow$)      & 43.18 & 44.55 & 42.47 & \underline{44.76} & \textbf{45.17} && 37.00 & 38.61 & 33.86 & \underline{39.29} & \textbf{39.59} \\
    mAR@L  & ($\uparrow$)     & 53.85 & 55.70 & 52.14 & \underline{56.07} & \textbf{56.52} && 54.03 & 57.42 & 48.66 & \underline{58.13} & \textbf{58.55} \\
    \cmidrule(lr){1-7}
    \cmidrule(lr){9-13}
    \end{tabular}
    \\\footnotesize{$^{*}$We report the suppression results only after QUBO, not including soft-scoring in QSQS. $^{\dag}$Proposed method}
    \caption{The experimental results. The best values are shown in \textbf{bold}, and the second-best values are \underline{underlined}. ``S'', ``M'', and ``L'' denote the size of bounding boxes such as small, medium, and large.}
    \label{tab:results_faster_rcnn}
\end{table*}

\vskip.5\baselineskip\noindent\textbf{CrowdHuman dataset. }
The right part of \cref{tab:results_faster_rcnn} shows the results on the CrowdHuman dataset. The proposed methods, QAQS and QAQS-C, outperform QSQS by 4.39 and 4.54 points in mAP and 9.47 and 9.89 points in mAR, respectively.
Like the COCO dataset, the proposed methods significantly improve mAP and mAR for large objects (mAP@L and mAR@L).
However, the degree of improvement is larger than that for the COCO dataset.
This indicates that the proposed methods are more effective for crowded scenes where (partial) occlusion is likely to occur.

\begin{figure*}[t]
     \centering
     \begin{subfigure}[b]{0.71\linewidth}
         \centering
         \includegraphics[width=\linewidth]{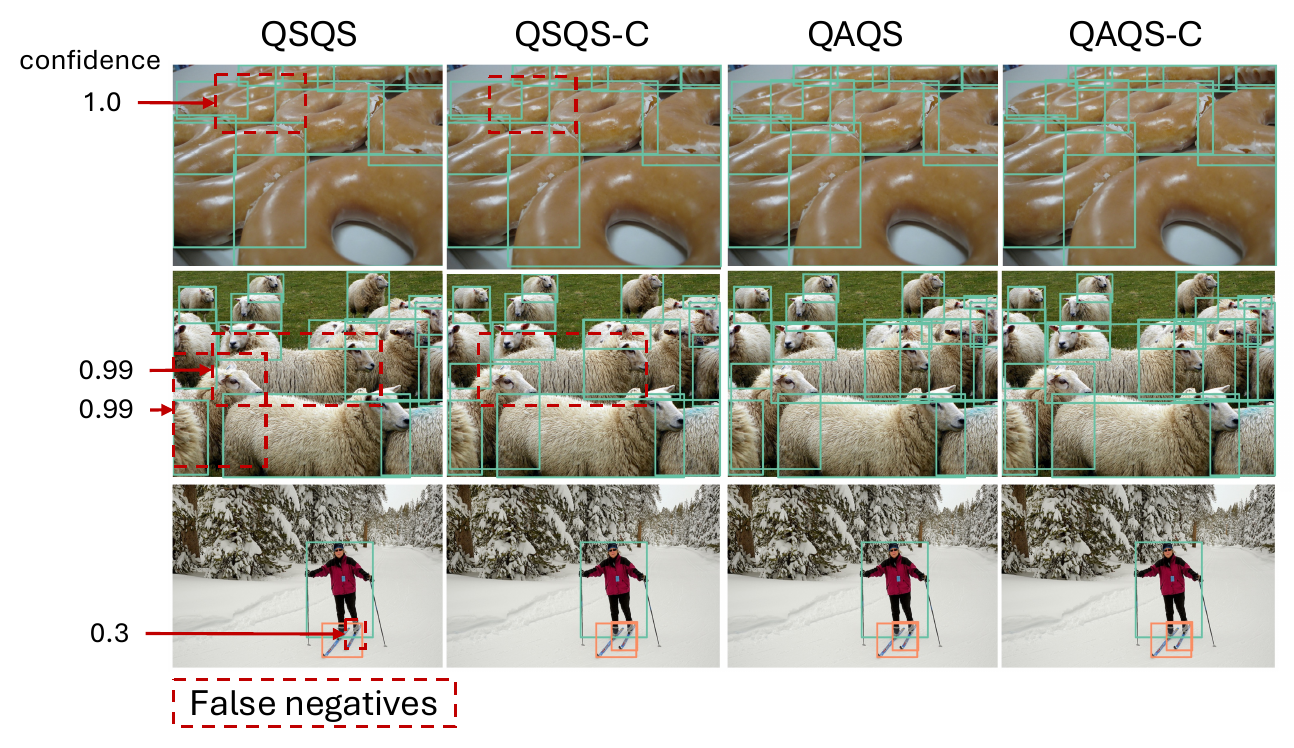}
         \caption{Qualitative results}
         \label{fig:qualitative}
     \end{subfigure}
     \begin{subfigure}[b]{0.22\linewidth}
         \centering
         \includegraphics[width=\linewidth]{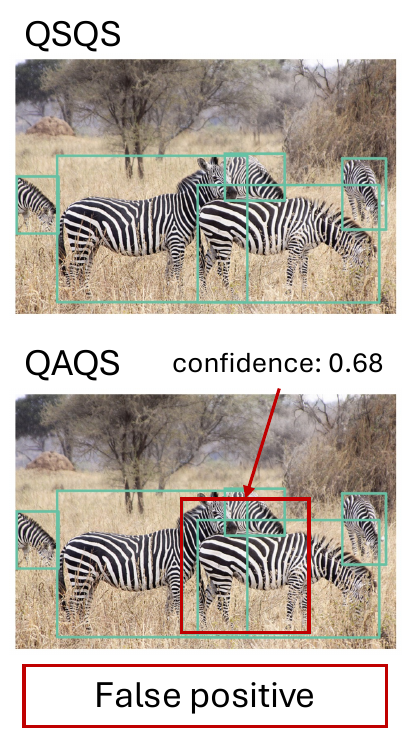}
         \caption{Potential drawbacks}
         \label{fig:potential_drawback}
     \end{subfigure}
        \caption{Visualization of suppressed predictions. Confidence scores of false negatives (\cref{fig:qualitative}) and false positives (\cref{fig:potential_drawback}) are shown outside of each picture.}
        \label{fig:visualization}
\end{figure*}

\vskip.5\baselineskip\noindent\textbf{Qualitative results. }
\Cref{fig:visualization} visualizes the detection results after QUBO-based suppressions.
As shown in \cref{fig:qualitative}, the proposed methods can detect the (partially-) occluded objects suppressed by the existing method, QSQS.
The objects suppressed by QSQS overlap with multiple predictions, leading QSQS to impose a larger penalty on these predictions. This excessive penalty can be appropriately mitigated by incorporating appearance features.
The drawback of introducing an appearance feature is shown in \cref{fig:potential_drawback}. 
A prediction that partially contains multiple objects has a different appearance from a prediction that mainly represents a single object. The appearance feature may lead to over-detection, as it limits the penalty for such predictions.

%% file: sec/experiments/qaqs.tex
\subsection{Ablation study of coefficient matrix}
We investigate the individual contribution of two features newly introduced to the $Q$ matrix through an ablation study.
\Cref{tab:ablation_qubo} show the results for the COCO and CrowdHuman datasets.
QSQS-C denotes the formulation where the pairwise score of QSQS $(w_2P_1+w_3P_2)$ is replaced with $\bm{v}(w_2 P_1+w_3 P_2)\bm{v}^\top$.
The results show that both features improve performance compared to baseline QSQS, and the appearance feature is mainly affected.
\setlength\tabcolsep{0.75mm} 
\begin{table}[tbh]
    \centering
    \small
    \begin{tabular}{l ccc ccc}
        \toprule
        & \multicolumn{3}{c}{COCO 2017}& \multicolumn{3}{c}{CrowdHuman}\\
         \cmidrule(lr){1-4}
         \cmidrule(lr){5-7}
         Method & Time$^*$ & mAP & mAR@100 & Time$^*$ & mAP & mAR@100 \\
         \cmidrule(lr){1-4}
         \cmidrule(lr){5-7}
         QSQS   & 21 & 34.03 & 42.90 & 32 & 31.23 & 36.19 \\ 
         QSQS-C & 21 & 34.50 & 44.33 & 30 & 33.10 & 39.19 \\ 
         QAQS   & 24 & 35.25 & 45.50 & 37 & 35.62 & 42.57 \\ 
         QAQS-C & 24 & 35.35 & 45.90 & 35 & 35.77 & 42.91 \\ 
        \bottomrule
        \multicolumn{7}{l}{\footnotesize{$^*$ Runtime for suppression. Milliseconds per image [ms/image]}}
    \end{tabular}
    \caption{Ablation study of QUBO-based suppression. }
    \label{tab:ablation_qubo}
\end{table}

\begin{figure}
    \centering
    \includegraphics[width=0.8\linewidth]{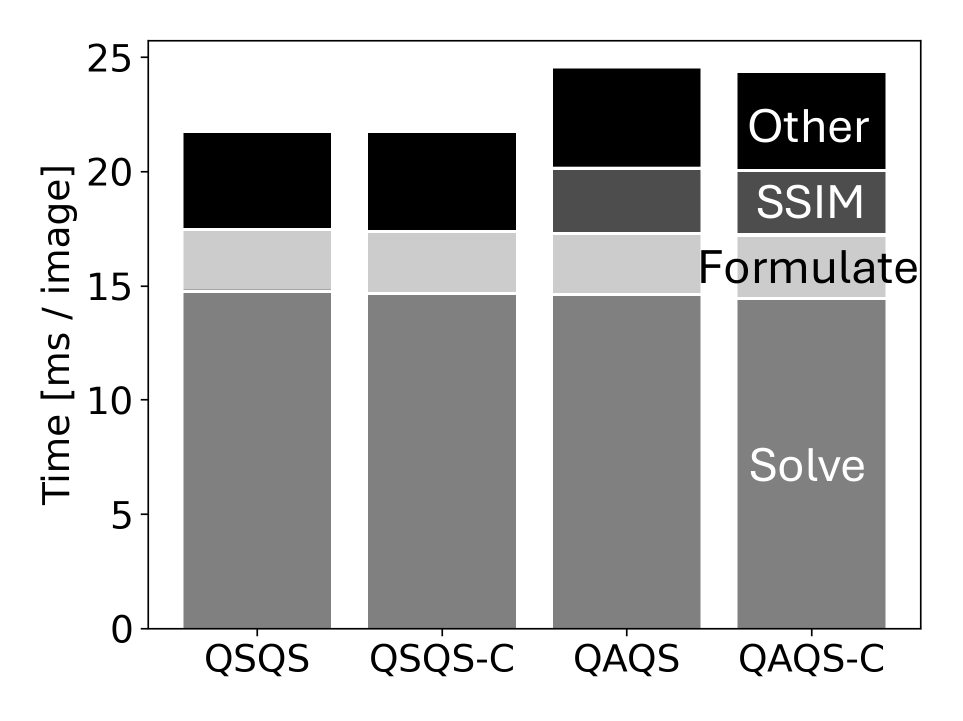}
    \caption{Breakdown of QUBO-based suppression runtime per image of COCO dataset shown in \cref{tab:ablation_qubo}. Results on the CrowdHuman dataset show a similar tendency.}
    \label{fig:breakdown_coco}
\end{figure}

The total runtime of QAQS and QAQS-C is slightly longer than that of QSQS and QSQS-C due to the appearance feature calculation. However, the difference is 2-6~ms per image, which is a negligible overhead compared to the total runtime.
The frames-per-seconds (fps) is calculated as the reciprocal of the total detection runtime per image that equals the sum of forward and suppression time per image.
For the COCO dataset, the forward time for Faster R-CNN without suppression is around 54~ms per image, and the runtime of our QAQS-C is approximately 24~ms per image. 
Thus, the overall detection throughput is approximately $1000\times\frac{1}{54+24} \fallingdotseq 12.8$ fps.
\Cref{fig:breakdown_coco} shows the runtime breakdown for QUBO-based suppressions. The current major bottleneck is the runtime of the QUBO solver, which spends approximately half of the runtime. If the solver runtime can be reduced from 14~ms to 1~ms by replacing the QUBO solver with a faster quantum computer in the future, the overall throughput will improve to approximately 18.2~fps.
10-20~fps is sufficiently fast for some industrial applications, such as surveillance inside the factory, because objects do not move fast in these situations.

%% file: sec/experiments/ssim.tex
\subsection{Ablation study of our implementation of SSIM}
\begin{figure}[t]
     \centering
     \begin{subfigure}[b]{0.45\linewidth}
         \centering
         \includegraphics[width=\linewidth]{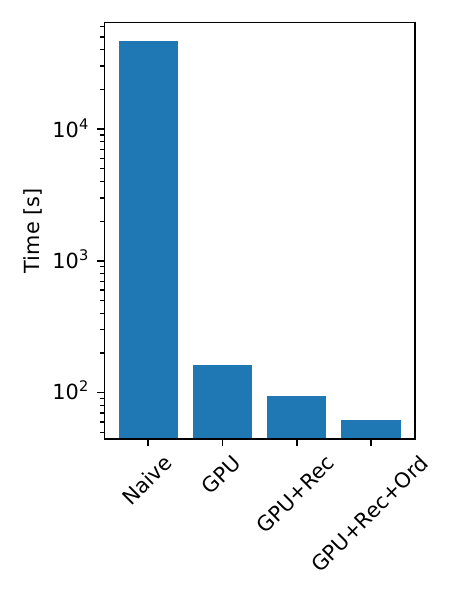}
         \caption{Computation time}
         \label{fig:ssim_time}
     \end{subfigure}
     \hfill
     \begin{subfigure}[b]{0.45\linewidth}
         \centering
         \includegraphics[width=\linewidth]{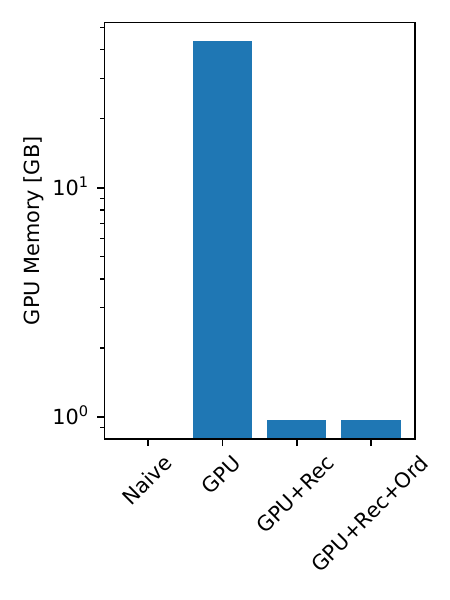}
         \caption{GPU memory usage}
         \label{fig:ssim_memory}
     \end{subfigure}
        \caption{Ablation study of SSIM computation. \textit{Naive} is the sequential computation on the CPU based on the implementation of scikit-image. \textit{GPU} represents the GPU parallelization. \textit{Rec} represents the recursive computation using the divide-and-conquer algorithm. \textit{Ord} shows the reordering of \textit{Intersection Matrix} to avoid redundant computation.}
        \label{fig:ssim_ablation}
\end{figure}
We investigate the impact of each technique in our SSIM implementation on the computation time and peak GPU memory usage.
\Cref{fig:ssim_ablation} shows the results for the CrowdHuman dataset, which has more pronounced benefits from our faster implementation.
We use the sequential computation of SSIM implemented on scikit-image
\footnote{\url{https://scikit-image.org}} 
as the baseline.
To summarize the experiments, applying all the techniques reduces the computation time from the baseline of approximately 2.0 and 10.5~s/image to 6 and 14~ms/image, respectively, at the cost of under 1~GB GPU memory usage.

\vskip.5\baselineskip\noindent\textbf{Effect of parallelization on GPU. }
We validate the efficacy of GPU parallelization by comparing it to baseline implementation. We do not consider GPU memory usage because the baseline only uses CPU.
Although the baseline requires 9,866 and 46,178 seconds to compute SSIM for all COCO and CrowdHuman validation images, the GPU-parallelized version only takes 43 and 163 seconds, respectively.
This parallelization significantly enhances the throughput, reducing the compute time from approximately 2 and 11~s/image to 8.6 and 37~ms/image for the COCO and CrowdHuman datasets.

\vskip.5\baselineskip\noindent\textbf{Effect of divide-and-conquer algorithm on GPU memory usage. }
We examine the impact of our divide-and-conquer algorithm on memory usage and calculation time by comparing GPU parallelization with and without divide-and-conquer.
By introducing the divide-and-conquer algorithm, the execution time is reduced from 43 and 163 seconds to 33 and 95 seconds on the COCO and CroudHuman datasets, respectively.
GPU memory usage is also significantly reduced from 
24 and 41~GB to 0.83 and 0.92~GB for the COCO and CrowdHuman datasets.

\vskip.5\baselineskip\noindent\textbf{Effect of the omission of redundant computations. }
The runtime is further reduced with no extra GPU usage by omitting the redundant computation.
We demonstrate this by comparing divide-and-conquer+GPU parallelization with and without omitting redundant computation.
By omitting redundant computation, the computation time is reduced from 33 and 95 seconds to 30 and 62 seconds for the COCO and CrowdHuman datasets without any increase in GPU memory usage.

%% file: sec/discussion.tex
\subsection{One reason for higher performance of the proposed formulations}
The higher mAP and mAR of proposed methods than QSQS indicate that they are less likely to suppress true positives. We can mathematically explain the reason for this behavior.
The non-diagonal elements of the coefficient matrix represent the penalty among predictions. If the penalty among predictions is low, the predictions are less likely to be suppressed.
The absolute value of non-diagonal elements of the QAQS coefficient matrix is smaller than that of QSQS because SSIM is less than or equal to 1.
Similarly, the absolute value of non-diagonal elements of the QAQS-C coefficient matrix is smaller than that of QAQS because the confidence score is between 0 and 1.
This analysis indicates that QSQS imposes a more severe penalty than the penalties applied by QAQS and QAQS-C.
Therefore, the proposed formulations, QAQS and QAQS-C, are less likely to suppress true positives than QSQS.


\subsection{Guidelines for further acceleration}
By improving the sparsity of the coefficient matrix, the runtime for SSIM and optimization calculation might be reduced. The high sparsity indicates that the number of pairs for SSIM computation is small. Thus, we can simply reduce the number of SSIM computations. Additionally, the QUBO may be efficiently solved by exploiting the sparsity of the coefficient matrix~\cite{Rehfeldt2023}.

To improve the sparsity of the coefficient matrix, we replace IoU with Generalized IoU (GIoU)~\cite{RezatofighiTGS019GIoU}. GIoU takes a smaller value than IoU by definition. Considering the consistency with the case of IoU, the pairwise score ($P_1$ and $P_2$) of the pairs with negative GIoU values is set to 0. With this modification, the runtime of QAQS and QAQS-C is almost halved, maintaining the performance. The runtime of QSQS is also halved with a slight degradation. It will depend largely on trial and error whether the sparsity of the coefficient matrix is improved without affecting the optimal solution, but this method would be a promising heuristic to achieve it.
See Supplementary C for more details.

%% file: sec/related_work.tex
This section reviews the related studies from three topics: object detection, image similarity metric, and application of quantum computing for computer vision.

\vskip.5\baselineskip\noindent\textbf{Object detection. }
Non-Maximum Suppression (NMS) is applied to many object detection models to suppress redundant predictions.
NMS iteratively chooses a prediction in descending order of confidence score and removes predictions whose IoU is higher than the threshold.
This procedure causes a well-known drawback of NMS, such as high sensitivity of the threshold and missing objects in crowded scenes where many objects are likely to be occluded.
To address these issues, the model architectures~\cite{Chen2022DATE,YOLOV10,Lv2024detr} and suppression algorithms~\cite{bodla2017snms,ShenJXLK22CPCluster,shepley2023confluence} have been studied.
The recent suppression algorithms are mainly categorized into NMS variants~\cite{bodla2017snms,zhou2017cad,Liu2019AdaNMS,He2019SofterNMS,Nils2020vgnms,Huang2020VFGNMS,solovyev2021weighted,shepley2023confluence}, graph-based methods~\cite{ShenJXLK22CPCluster}, and optimization-based methods~\cite{rujikietgumjorn2013qubo,lee2016individual,li2020qsqs}.
The most famous NMS variant is Soft-NMS~\cite{bodla2017snms}, which uses a scoring-based penalty for overlapped predictions instead of threshold-based suppression.
CP-Cluster~\cite{ShenJXLK22CPCluster} is a graph-based method that treats the predictions as a graph whose adjacent matrix is based on the overlap among predictions. 
The main focus of our study is optimization-based suppression, which formulates the suppression as an optimization problem. The redundant predictions are suppressed by optimizing the objective function that considers the confidence score and pairwise score of predictions.
Our primary goal is to foster promising applications for quantum computers by refining the QUBO-based suppression algorithm rather than developing an ultimate suppression algorithm that outperforms all other suppression algorithms in every possible scenario.

\vskip.5\baselineskip\noindent\textbf{Image similarity metric. }
Towards similar image retrieval and image quality evaluation, methods to quantify the similarity between images have been developed. The simplest method is to compare the color histograms of images computed from the frequency of pixel values. Considering its application to QUBO, its value range $[0, 1]$ is preferable to adjust the scale of the coefficient matrix but computationally expensive.
Keypoint-based methods~\cite{lowe1999sift,alcantarilla2013akaze} are computationally reasonable but do not work well for small images.
SSIM~\cite{wang2004ssim,hore2010psnr_ssim} and Peak Signal to Noise Ratio (PSNR)~\cite{hore2010psnr_ssim} are designed to reflect human perception and are mainly used for image quality evaluation. SSIM is also used as the loss function of generative models~\cite{ZhaoGFK17ImgRestore}. In contrast to SSIM, PSNR is unsuitable for integration into QUBO because its value range is not scaled to $[0, 1]$.
Deep learning-based methods have also been studied. 
\citet{lee2016individual} used CNN features for optimization-based suppression, but this approach is model-specific and computationally expensive.
Learned Perceptual Image Patch Similarity (LPIPS)~\cite{zhang2018lpips} quantifies the visual similarity of images based on the feature maps of a pre-trained image classifier. LPIPS is closer to human perceptions but computationally expensive than conventional metrics directly calculated from pixel values.

\vskip.5\baselineskip\noindent\textbf{Application of quantum computing for computer vision. }
Quantum computing has recently been applied to computer vision tasks that involve combinatorial optimization problems~\cite{o2018nonnegative,golyanik2020quantum,birdal2021quantum,doan2022}.
In multi-object tracking~\cite{Ngeni2023,zaech2022}, the detection of the same object between consecutive video frames should be matched. Quantum computer solves the QUBO formulation for this matching problem.
\citet{meli2022} approximates the estimation problem of momentum transformation from point crowd as QUBO. The momentum transformation is estimated by iteratively solving the approximated QUBO.
Although these applications are formulated as constrained problems, the QUBO-based suppression is unconstrained.

%% file: sec/conclusion.tex
We propose new formulations of QUBO-based suppression based on the hypothesis that distinguishing predictions for occluded objects from redundant predictions enhances object detection performance.
Our formulations integrate the SSIM and the product of confidence scores between predictions into the coefficient matrix to differentiate occlusion from redundancy.
The proposed methods outperform the SOTA QUBO-based suppression, achieving up to 4.54 points improvement in mAP and up to 9.89 points gain in mAR for crowded and non-crowded datasets.
The proposed methods show a significant performance improvement for the crowded dataset. These results empirically validate our hypothesis.
The experiments are conducted with a classical solver, but our software is compatible with quantum solvers. Our quantum-ready methods will be more effective and beneficial with the advent of practical quantum computers.